\documentclass[10pt,twocolumn,letterpaper]{article}

\usepackage[pagenumbers]{cvpr} %

\usepackage{graphicx}
\usepackage{amsmath}
\usepackage{amssymb}
\usepackage{booktabs}
\usepackage{color}
\usepackage{multirow}
\usepackage{makecell}
\usepackage{overpic}
\usepackage{subcaption}
\usepackage{caption}

\makeatletter
    \newcommand{\vast}{\bBigg@{3}}
    \newcommand{\Vast}{\bBigg@{3.5}}
    \newcommand{\vastt}{\bBigg@{4}}
    \newcommand{\Vastt}{\bBigg@{4.5}}
\makeatother

\definecolor{darkgreen}{RGB}{30,150,30}
\definecolor{darkblue}{RGB}{0,0,127}
\definecolor{darkyellow}{RGB}{171,133,0}
\definecolor{darkred}{RGB}{180,20,20}
\definecolor{darkmagenta}{RGB}{200,0,200}
\definecolor{darkcyan}{RGB}{0,127,127}

\newif\ifdrafting 
\draftingfalse %

\ifdrafting
  \newcommand{\GL} [1] {\textcolor{darkgreen}{[GL: #1]}}
  \newcommand{\VJ} [1] {\textcolor{darkblue}{[VJ: #1]}}
  \newcommand{\ls} [1] {\textcolor{darkmagenta}{[LS: #1]}}
  \newcommand{\ds} [1] {\textcolor{red}{[DS: #1]}}
  \newcommand{\TODO} [1] {{\color{darkcyan}{\bf [TODO: #1]}}}
  
\else
  \newcommand{\GL} [1] {}
  \newcommand{\VJ} [1] {}
  \newcommand{\ls} [1] {}
  \newcommand{\ds} [1] {}  
  \newcommand{\TODO} [1] {}
  
\fi

\newcommand{\method}[1]{LOCATE}
\newcommand{\s}[1]{PartSelect}

\usepackage[pagebackref,breaklinks,colorlinks]{hyperref}

\usepackage[capitalize]{cleveref}
\crefname{section}{Sec.}{Secs.}
\Crefname{section}{Section}{Sections}
\Crefname{table}{Table}{Tables}
\crefname{table}{Tab.}{Tabs.}

\def\cvprPaperID{575} %
\def\confName{CVPR}
\def\confYear{2023}

\begin{document}

\title{LOCATE: Localize and Transfer \\ Object Parts for Weakly Supervised Affordance Grounding}
\author{Gen Li\textsuperscript{1}\qquad Varun Jampani\textsuperscript{2}\qquad
Deqing Sun\textsuperscript{2}\qquad
Laura Sevilla-Lara\textsuperscript{1}\\ \\
\textsuperscript{1}University of Edinburgh\quad
\textsuperscript{2}Google Research\\
}

\twocolumn[{%
\renewcommand\twocolumn[1][]{#1}%
\maketitle

\begin{center}
    \centering
    \captionsetup{type=figure}
    \begin{overpic}[width=.95\linewidth]{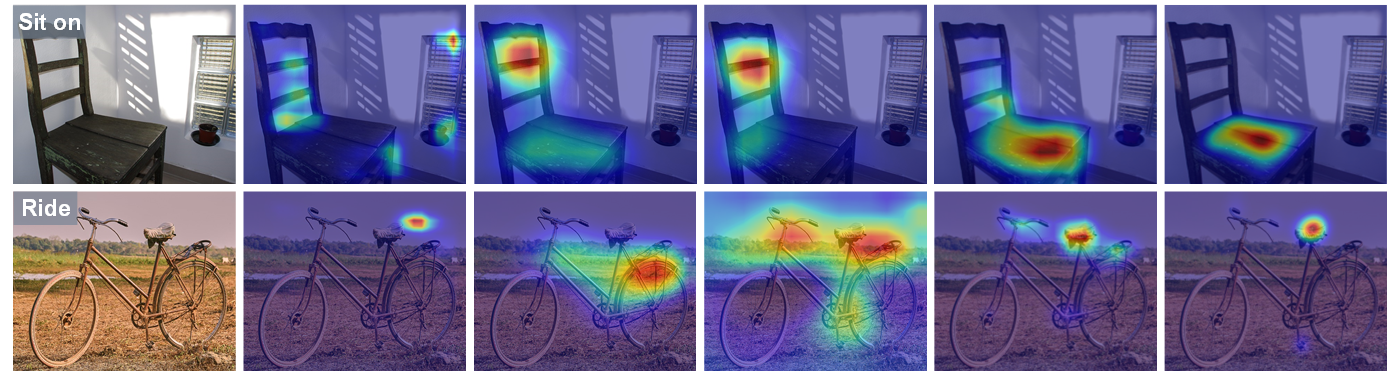}
    \put(4.6,-1.7){{\footnotesize{Object Image}}}
    \put(20.9,-1.7){{\footnotesize{Hotspots\cite{grounded}}}}
    \put(35.2,-1.7){{\footnotesize{Cross-view-AG\cite{ag_from_exocentric_imgs}}}}
    \put(51.2,-1.7){{\footnotesize{Cross-view-AG+\cite{ag_from_exocentric_imgs+}}}}
    \put(69.6,-1.7){{\footnotesize{\method~ (Ours)}}}
    \put(90.4,-1.7){{\footnotesize{GT}}}
    \end{overpic}
    \vspace*{.5em}
    \captionof{figure}{State-of-the-art methods in weakly supervised affordance grounding often fail to make accurate predictions for objects with complex structures, e.g., chairs and bicycles.
    To address this, our model (\method~) focuses on localizing and transferring features of object parts, which is able to produce more accurate results.
      } %
    \label{fig:banner}
\end{center}%
}]

\begin{abstract}
\vspace*{-.55em}
Humans excel at acquiring knowledge through observation.
For example, we can learn to use new tools by watching demonstrations. This skill is fundamental for intelligent systems to interact with the world. A key step to acquire this skill is to identify what part of the object affords each action, which is called affordance grounding. 
In this paper, we address this problem and propose a framework called LOCATE that can identify matching object parts across images, to transfer knowledge from images where an object is being used (exocentric images used for learning), to images where the object is inactive (egocentric ones used to test). 
To this end, we first find interaction areas and extract their feature embeddings.
Then we learn to aggregate the embeddings into compact prototypes (human, object part, and background), and select the one representing the object part.
Finally, we use the selected prototype to guide affordance grounding. 
We do this in a weakly supervised manner, learning only from image-level affordance and object labels.
Extensive experiments demonstrate that our approach outperforms state-of-the-art methods by a large margin on both seen and unseen objects.\footnote{Project page: \url{https://reagan1311.github.io/locate}.}
\end{abstract}

\section{Introduction}
A fundamental skill of humans is learning to interact with objects just by observing someone else performing those interactions \cite{neural_observation_learning}.
For instance, even if we have never played tennis, we can easily learn where to hold the racket just by looking at a single or few photographs of those interactions. 
Such learning capabilities are essential for intelligent agents to understand what actions can be performed on a given object.
Current visual systems often focus primarily on recognizing {\em what} objects are in the scene (passive perception), rather than on {\em how} to use objects to achieve certain functions (active interaction).
To this end, a growing number of studies \cite{affordancenet, aff_spcae, aff_semantic_relations, aff_detect_task_specific_grasp} have begun to utilize affordance \cite{gibson} as a medium to bridge the gap between passive perception and active interaction.
In computer vision and robotics~\cite{aff_robot_survey, aff_vis_survey}, affordance typically refers to regions of an object that are available to perform a specific action, e.g., a knife handle affords holding, and its blade affords cutting. 

\begin{figure}[!t]
\centering
    \includegraphics[width=1\linewidth]{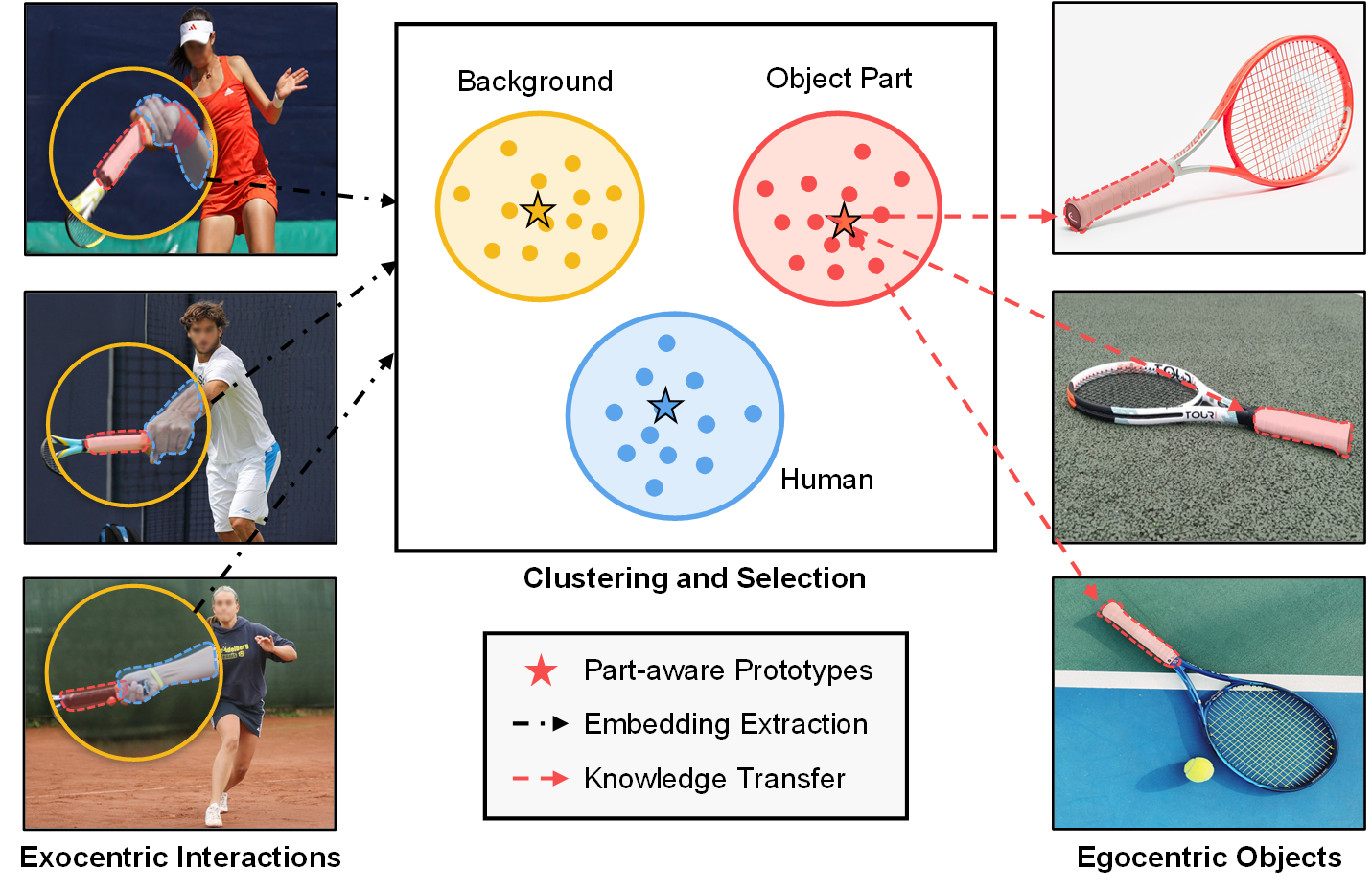}
    \caption{General illustration of \method~. We first extract the embeddings under the region of interest where the exocentric interactions are happening, and then split these embeddings into several clusters. In the end, the prototype of the object-part cluster is selected to supervise the egocentric affordance grounding.}
   \label{fig:teaser}
\end{figure}

In this paper, we focus on the task of affordance grounding, i.e, locating the object regions used for a given action.
Previous methods \cite{demo2vec, affordancenet, aff_with_CNN_umd, tool_parts_iff, learn2act_properly} have often treated affordance grounding as a fully supervised semantic segmentation task, which requires costly pixel-level annotations.
Instead, we follow the more realistic setting~\cite{grounded, ag_from_exocentric_imgs,ag_from_exocentric_imgs+} where the task is learning object affordances by observing human-object interaction images. That is, given some interaction images, such as those in Fig.~\ref{fig:teaser}, along with the corresponding label (e.g., ``hold"), the aim is to learn affordance grounding on the novel instances of that object. This is a weakly-supervised
problem setting where only the image-level labels are given without any per-pixel annotations.
Concretely, given several third-person human-object interaction images (exocentric) and one target object image (egocentric), our goal is to extract affordance knowledge and cues from exocentric interactions, and perform affordance grounding in the egocentric view by using only affordance labels.

There are several key challenges underlying the problem of affordance grounding.
The first is due to the nature of the supervision, where only image-level affordance labels are given, being a weakly supervised problem. Here, the system needs to automatically reason about affordance regions just from classification labels.
Second, human-object interactions often introduce heavy occlusion of object parts by interacting humans.
In other words, the object part that the system needs to predict for a particular affordance (e.g., a mug handle for the ``holding" affordance) in an exocentric image can often be the part that is occluded (e.g., by hands). 
Third, interactions are of great diversity. The way humans interact with objects varies across individuals resulting in diverse egocentric interaction images.
Lastly, there is a clear domain gap between exocentric and egocentric images where the former have clutter, occlusion etc., and the latter are cleaner (e.g., in Fig.~\ref{fig:teaser}). This makes affordance knowledge transfer particularly challenging.

In this work, we propose a framework called \method~ that addresses these core challenges by locating the exact object parts involved in the interaction from exocentric images and transferring this knowledge to inactive egocentric images. 
Refer to Fig.~\ref{fig:teaser} for the illustration. Specifically, we first use the class activation mapping (CAM) \cite{cam} technique to find the regions of human-object-interaction in exocentric images.
Despite being trained for the interaction recognition
task, we observe that CAM can generate good localization maps for interaction regions.
We then segment this region of interest further into regions corresponding to human, object part, and background. We do this by extracting embeddings and performing k-means clustering to obtain several compact prototypes. %
Next, we automatically predict which of these prototypes corresponds to the object part relevant to the affordance.
To this end, we propose a module named \s~ that leverages part-aware features and attention maps from a self-supervised vision transformer (DINO-ViT \cite{dino-vit}) to obtain the desired prototype.
Finally, we use the object-part prototype as a high-level pseudo supervision to guide egocentric affordance grounding.

Our contributions can be summarized as follows.
(1) We propose a framework called \method~ that extracts affordance knowledge from weakly supervised exocentric human-object interactions, and transfers this knowledge to the egocentric image in a localized manner.
(2) We introduce a novel module termed \s~ to pick affordance-specific cues from human-object interactions. The extracted information is then used as explicit supervision to guide affordance grounding on egocentric images. 
(3) LOCATE achieves state-of-the-art results with far fewer parameters and faster inference speed than previous methods, and is able to locate accurate affordance region for unseen objects. See Fig.~\ref{fig:banner} for examples of our results and comparison to state-of-the-art.

\section{Related Work}

\noindent\textbf{Visual Affordance Learning.}
Affordances are actively studied in robotics and computer vision due to their great potential to bridge the gap between perception and action.
In general, affordance-related research in computer vision aims to localize objects or object parts where specific actions can be performed.
Earlier work \cite{cad120, affordancenet, learn2act_properly, aff_obj_parts} mainly focused on the fully supervised setting.
Due to the scarcity of affordance datasets, this work normally required costly manual annotations to construct large-scale datasets \cite{tool_parts_iff, aff_with_CNN_umd, weakly_supervised_affordance_detection, learn2act_properly}.
To alleviate the annotation and sensing requirement, new research~\cite{grounded, weakly_supervised_affordance_detection, weakaff2, ag_from_exocentric_imgs} explored acquiring affordance information in the weakly supervised setting.
In particular, Sawatzky et al. \cite{weakly_supervised_affordance_detection} proposed a convolutional network to tackle affordance detection using only a few keypoint annotations, and Nagarajan et al. \cite{grounded} introduced a method to infer the object affordance regions by directly learning from human-object interaction videos.
In this work, we also focus on learning affordances from weak supervision, i.e., several human-object interaction images and corresponding image-level affordance labels.

\begin{figure*}[ht]
\centering
   \includegraphics[width=0.95\linewidth]{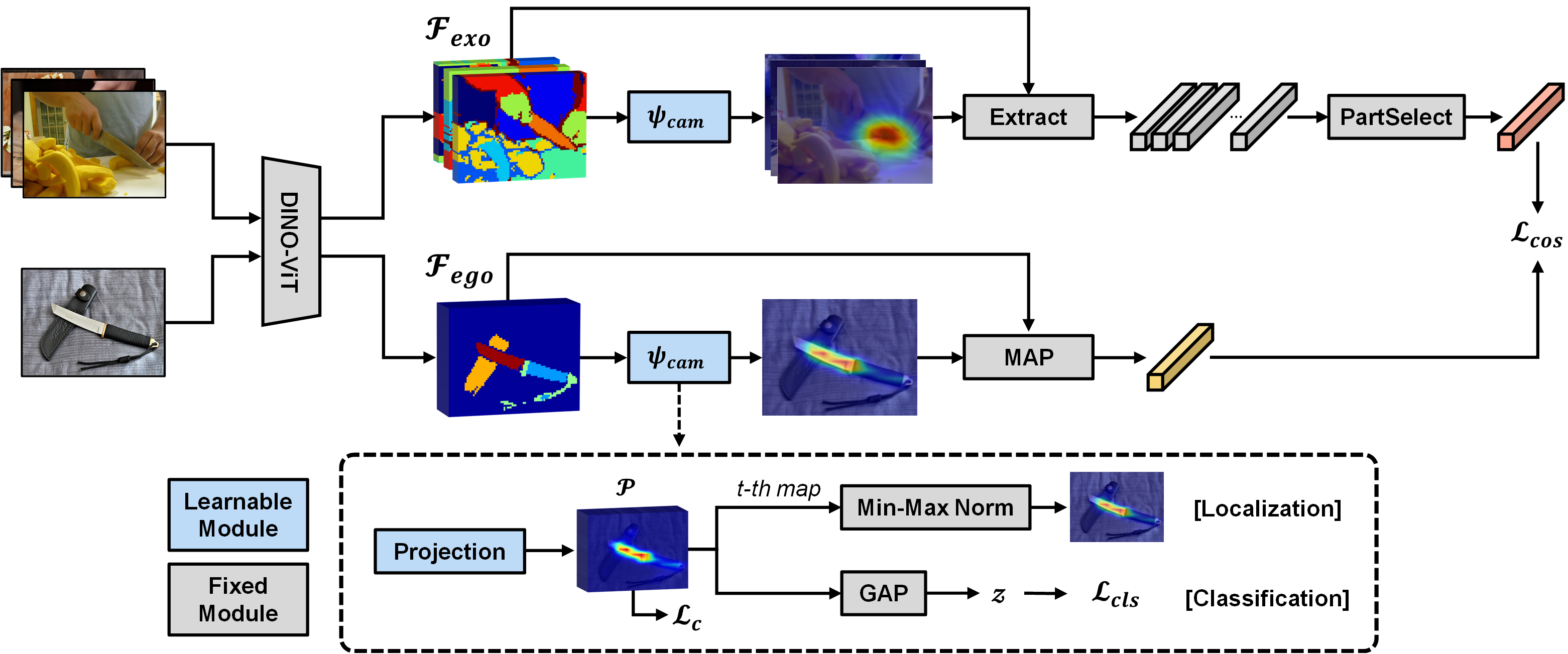}
   \caption{Overview of the proposed \method~ framework. It achieves part-level knowledge transfer in three steps: 1) locating interaction regions with $\psi_{cam}$ (Sec.~\ref{sec:3.1}), 2) object-part embedding selection with \s~ (Sec.~\ref{sec:3.2}), and 3) part-level knowledge transfer with $L_{cos}$ (Sec.~\ref{sec:3.3}). Details for \s~ are shown in Fig.~\ref{fig:selector_all}. At test time, only the egocentric branch is maintained.}
   \label{fig:framework}
\end{figure*}

\vspace{1mm}
\noindent\textbf{Weakly Supervised Affordance Grounding.}
Unlike object localization that aims to find where the object is, in affordance grounding we focus on locating functional regions of objects, which is a more fine-grained localization task.
Existing weakly supervised object localization \cite{bridge_wsol, EIL, cream, TS-CAM, SPA, Acol} and semantic segmentation methods \cite{RCA, cllims, Re-CAM, self_wsss, railroad} are mainly based on class activation mapping (CAM) \cite{cam}, the seminal work that has greatly facilitated the development of weakly supervised learning.
Similarly, most affordance grounding research \cite{grounded, ag_from_demo_video, ag_from_exocentric_imgs} also adopted CAM or its variants \cite{grad_cam, grad_cam++} to generate activation maps as prediction.
However, previous work only used CAM in the inference stage, lacking proper supervision.
In contrast, to provide explicit guidance at the training stage, we utilize CAM to produce localization maps in the forward pass and supervise it in a prototypical learning fashion.

\vspace{1mm}
\noindent\textbf{Knowledge Distillation across Domains.}
Knowledge distillation \cite{kd} is a process of transferring knowledge from a network to another, where inputs normally come from the same domain.
Recently, many studies \cite{charades-ego, ego-exo, demo2vec, grounded, ag_from_exocentric_imgs} have begun to pay more attention to knowledge distillation in different domains.
This work transferred representations from the domain with numerous datasets to the other one that typically has low dataset scale and diversity, so that the model can benefit from the large quantity of data in the first domain.
For instance, Li et al. \cite{ego-exo} proposed a framework to learn egocentric video representations from large-scale third-person video datasets, and Luo et al \cite{ag_from_exocentric_imgs} devised a knowledge transfer architecture to transfer affordance-specific features from exocentric view to egocentric view.
In this paper, we follow the same setting as \cite{ag_from_exocentric_imgs}, using exocentric (third-person) human-object interaction images as the learning target.
The objective of this work is to observe where and how humans interact with an object when doing a specific action, and then transfer the knowledge onto the target object.

\vspace{1mm}
\noindent\textbf{Self-supervised Vision Transformer.}
Self-supervised Vision Transformers (ViT) \cite{vit}, such as DINO \cite{dino-vit}, MAE \cite{mae}, and BEiT \cite{beit}, have demonstrated immense potential in unsupervised dense prediction tasks.
Specifically, the features extracted from the self-attention layer exhibit the ability to separate different objects and generate reliable pseudo segmentation maps without requiring any manual annotations.
Building on this property, recent work \cite{dino-discover, dino-LOST, dino-spectral, dino-tokencut, dino-UnsupervisedSS, affcorrs} has achieved exceptional results in unsupervised segmentation and localization.
Notably, Amir et al. \cite{deepvit} demonstrated that the pretrained features extracted from DINO-ViT encode fine-grained semantic information, which can yield excellent performance for part co-segmentation and semantic correspondence.
Inspired by this work, we leverage DINO-ViT features to identify matching object parts between exocentric and egocentric images.

\section{Method}
Given several exocentric interaction images and one egocentric object image, our goal is to extract affordance-related knowledge from exocentric interactions, and transfer it to egocentric images so that the affordance region can be located even for an inactive object. 
During training, the only supervision available are image-level affordance labels.
In the inference stage, taking an egocentric image and an affordance label as input, the model needs to predict the corresponding affordance region.

The core idea of our approach is to exclude distracting information, e.g., human and background, when extracting affordance-specific features from the exocentric view, and perform fine-grained part-level knowledge transfer from exocentric images to egocentric ones.
To this end, we set up the framework \method~ to transfer the knowledge in three steps (See Fig.~\ref{fig:framework}).
First, we utilize CAM to generate localization maps for exocentric images, and extract corresponding feature embeddings with high activation in the localization maps (Sec.~\ref{sec:3.1}). 
Then, we propose \s~ that leverages part-aware deep features to remove irrelevant information while preserving embeddings that can represent affordance cues (Sec.~\ref{sec:3.2}).
Finally, we use the output from \s~ to supervise the egocentric affordance grounding in an explicit manner (Sec.~\ref{sec:3.3}).

\subsection{Locating Interaction Regions}

\label{sec:3.1}
To determine where neural networks focus on for recognition, we adopt the technique of CAM \cite{cam} to generate class-aware localization maps, which has been widely used in weakly supervised tasks.
The vanilla CAM generates localization maps as a post-processing step that cannot be guided during training.
However, our goal is to extract affordance-specific cues from exocentric images, and use these cues as explicit supervision for the egocentric view.
Therefore, in order to obtain localization maps during the training phase, we produce class-specific feature maps instead by adding a class-aware convolution layer, which has proven to be identical to the generation process in CAM \cite{Acol}.
Specifically, for input images $\{I_{exo}, I_{ego}\}$~($I_{exo}\!=\!\{I_1, I_2, ..., I_N\}$), 
we first extract deep features $\{\mathcal{F}_{exo}, \mathcal{F}_{ego}\}\in\mathbb{R}^{D\times H\times W}$ using a network $\phi$.
In our case, $\phi$ is a self-supervised vision transformer (DINO-ViT), whose features are part-aware and provide good part-level correspondences.
We then generate localization maps $\mathcal{P}$ and classification scores $\boldsymbol{z}$ as follows:
\begin{equation}
	\mathcal{P} = \psi_{cam}(\mathcal{F}) \in\mathbb{R}^{C\times H\times W}, \quad
	\boldsymbol{z}={\rm GAP} (\mathcal{P}) \in\mathbb{R}^{C},
\end{equation}
where $\psi_{cam}$ starts with a projection layer consisting of a feed-forward layer followed by two convolutions to finetune features for the HOI recognition task, i.e., recognizing actions shown in the exocentric images.
Then a $1\times 1$ class-aware convolution layer is added to yield localization maps, converting the number of channels to $C$, where $C$ denotes the number of total interaction categories.
Therefore, each map $\mathcal{P}_c \in\mathbb{R}^{H\times W}$ represents the network activation for the $c$-th interaction.
Next, $\mathcal{P}$ is fed to a global average pooling (GAP) layer to obtain classification scores $\boldsymbol{z}$, which are used to calculate cross-entropy loss $\mathcal{L}_{cls}$ for optimization.

We notice that localization maps for the exocentric branch $\mathcal{P}_{exo}$ concentrate on the interaction areas, i.e, where the action takes place.
Since interaction areas carry strong affordance information, we therefore aim to collect embeddings from the high activation regions in exocentric localization maps.
Specifically, we first extract the localization map corresponding to ground-truth class, and conduct min-max normalization to constrain activation values to $[0, 1]$.
After that, we set a threshold $\tau$ to control the number of extracted embeddings, therefore embeddings with activation value greater than $\tau$ in the localization map will be extracted from deep features.
For multiple exocentric images, embeddings are extracted separately to produce $f_{exo}\!=\![f_1, ..., f_N]$, each $f_n$ containing a different number of embeddings.
All embeddings are then concatenated together $f_{exo} \in\mathbb{R}^{L\times D}$, where $L$ denotes the number of embeddings.

\subsection{Object-Part Embedding Selection}
\label{sec:3.2}

\begin{figure}
\centering
    \begin{subfigure}[b]{\linewidth}
    \centering
        \includegraphics[width=0.85\linewidth]{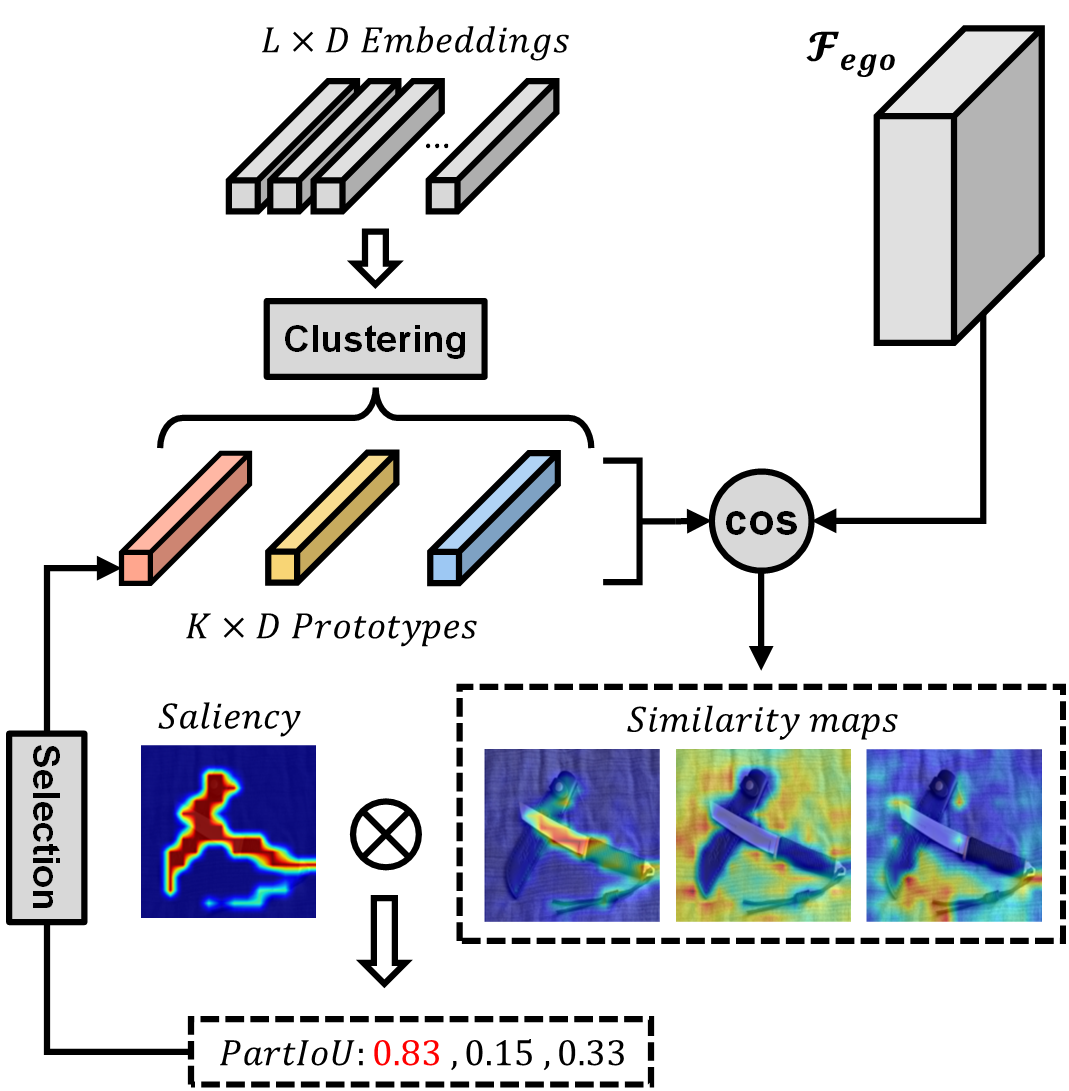}
        \vspace{1mm}
        \caption{Illustration of \s~.}
        \label{fig:selector}
        \vspace{2mm}
    \end{subfigure}
    
    \begin{subfigure}[b]{\linewidth}
    \centering
    \begin{overpic}[width=1.0\linewidth]{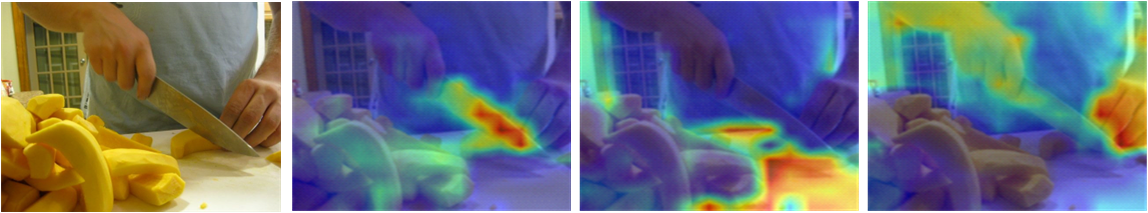}
    \put(7,-4){{\footnotesize{Exo-img}}}
    \put(30,-4){{\footnotesize{Object Part}}}
    \put(55,-4){{\footnotesize{Background}}}
    \put(83,-4){{\footnotesize{Human}}}
    \end{overpic}
    \vspace{1mm}
    \caption{Exocentric similarity maps.}
    \label{fig:exo_sim}
    \end{subfigure}
\caption{(a) \s~ picks the object-part prototype through clustering and selection, where $\bigotimes$ denotes the calculation for PartIoU score. (b) The similarity maps between prototypes and exocentric features confirm our statement that each prototype represents the object part, background, and human. 
\label{fig:selector_all}
}
\end{figure}

In general, interaction areas are composed of human, object part, and background. Our objective is to eliminate the interference information, and purely deliver embeddings representing the object part to guide the egocentric branch.
In consequence, we design \s~ to choose affordance-related embeddings from exocentric branch.
\s~ is illustrated in Fig.~\ref{fig:selector}. We first perform k-means clustering to get $K$ compact prototypes $\boldsymbol{p}\in\mathbb{R}^{K\times D}$ from extracted exocentric embeddings.
Next, we compute the cosine distance between each prototype and the egocentric deep features $\mathcal{F}_{ego}$ to get similarity maps $\mathcal{S}\in\mathbb{R}^{K\times H\times W}$:
\begin{equation}
S^{k,u,v}=\frac{\boldsymbol{p}_k \cdot  \mathcal{F}_{ego}^{u, v}}{\left\| \boldsymbol{p}_k \right\|\left\| \mathcal{F}_{ego}^{u, v} \right\|}.  
\end{equation}
Owing to the fine-grained semantic information of DINO-ViT deep features, embeddings of the same object parts bear high similarity.

To distinguish which prototype stands for the object part, we aggregate the self-attention maps from the last layer of the DINO-ViT to generate a saliency map $\mathcal{A} \in\mathbb{R}^{H\times W}$ for the egocentric image.
Given the saliency map and similarity maps, we introduce a metric $\gamma \in [0, 1]$ termed PartIoU to measure if a prototype carries object part information.  %
The PartIoU for the $k$-th prototype is defined as follows:

\begin{equation}
	\gamma = \frac{1}{2} \frac{\overline{\mathcal{S}_k}\cap \overline{\mathcal{A}}}{\overline{\mathcal{S}_k}} + \frac{1}{2} \frac{\overline{\mathcal{A}}}{\overline{\mathcal{S}_k}\cup \overline{\mathcal{A}}},
\end{equation}
where $\overline{\mathcal{S}_k}$, $\overline{\mathcal{A}} \in\{0, 1\}^{H\times W}$ are binary masks, we set the threshold as the average of each map to perform binarization.
The motivation of PartIoU is fairly straightforward, if $\overline{\mathcal{S}}$ belongs to a portion of $\overline{\mathcal{A}}$, then the intersection of $\overline{\mathcal{S}}$ and $\overline{\mathcal{A}}$ should equal $\overline{\mathcal{S}}$ itself, while the union of the two masks should be identical to $\overline{\mathcal{A}}$.
Finally, when the maximum PartIoU among $K$ prototypes is above a threshold $\mu$, \s~ will output the prototype with the largest PartIoU as the object-part representation. 
Otherwise, no prototype will be selected for the next step.
In Fig.~\ref{fig:exo_sim}, we visualize the similarity maps between prototypes and exocentric features to demonstrate that the extracted embeddings are clustered into human, object part, and background.

\subsection{Part-Level Knowledge Transfer}
\label{sec:3.3}
With the help of \s~, we find the prototype $f_{op}$ that represents the object part.
We then leverage it to perform supervision for egocentric localization maps $\mathcal{P}_{ego}$.
Concretely, we first perform masked average pooling (MAP) between the normalized localization map and extracted deep features to aggregate into one embedding:

\begin{equation}
	f_{ego} = \frac{\sum_{u=1, v=1}^{W, H} \mathcal{F}_{ego}^{u,v}\mathcal{P}_{ego}^{t,u,v}}{\sum_{i=1, j=1}^{W, H} \mathcal{P}_{ego}^{t,u,v}} \in\mathbb{R}^{D},
\end{equation}
where $t$ denotes the ground-truth category.
Then, a cosine embedding loss is applied to pull the embedding $f_{ego}$ towards the direction of $f_{op}$:
\begin{equation}
	\mathcal{L}_{cos} = \max(1 - \frac{f_{op} \cdot  f_{ego}}{\left\| f_{op} \right\|\left\| f_{ego} \right\|} - \alpha, 0), 
\end{equation}
as the two embeddings come from different domains, we thereby add $\alpha$ as a margin to compensate the domain gap.

In addition, since the affordance region typically denotes a portion of an object, we can thus impose a geometry loss to regulate its distribution.
Inspired by the co-part segmentation work \cite{scops}, we add a concentration loss to encourage egocentric localization maps to form a concentrated and connected component.
The concentration loss is formulated as
\begin{equation}
	\overline{u}_{c}=\sum_{u, v}u\cdot \mathcal{P}_{ego}^{c, u, v}/z_k, \,\,
	\overline{v}_{c}=\sum_{u, v}v\cdot \mathcal{P}_{ego}^{c, u, v}/z_k,
\end{equation}
\begin{equation}
	\mathcal{L}_{c} = \sum_{c} \sum_{u, v}\left\| \left< u, v\right> - \left<\overline{u}_{c}, \overline{v}_{c}\right>  \right\| \cdot \mathcal{P}_{ego}^{c, u, v}/z_c,
\end{equation}
where $\overline{u}_{c}$ and $\overline{v}_{c}$ represents the center of the $c$-th localization map along axis $u, v$, and $z_c=\sum_{u, v}\mathcal{P}_{ego}^{c, u, v}$ is a normalization term.
The concentration loss forces the high activation regions of the localization maps to be close to the geometric center.

Overall, we train the whole framework in an end-to-end manner, and use the following loss to optimize the model:

\begin{equation}
	\mathcal{L} = \mathcal{L}_{cls} + \lambda_{cos} \mathcal{L}_{cos} + \lambda_{c} \mathcal{L}_{c},
\end{equation}
where $\lambda_{cos}$, and $\lambda_{c}$ are loss weights that balance the three terms.
$\mathcal{L}_{cls}$ stands for the cross-entropy losses from the two branches.
At test time, only the ego branch is maintained, taking an affordance label $t$ and an egocentric image as input, the network extracts the $t$-th localization map as the prediction of affordance region.

\begin{table*}[!htb]
\centering
\begin{tabular}{cclccc@{\hskip 0.3in}ccc}
\toprule
\multicolumn{3}{c}{\multirow{2.5}{*}{\textbf{State-of-the-Art from Relevant Tasks}}} &  \multicolumn{3}{c@{\hskip 0.3in}}{\textbf{Seen}} & \multicolumn{3}{c@{\hskip 0.15in}}{\textbf{Unseen}} \\ \cmidrule(lr@{0.3in}){4-6} \cmidrule(r){7-9}
                                                     &                                          &                         & KLD$\downarrow$    & SIM$\uparrow$    & NSS$\uparrow$  & KLD$\downarrow$     & SIM$\uparrow$     & NSS$\uparrow$    \\ \midrule
\multirow{3}{*}{\makecell{Weakly Supervised \\ Object Localization*}} & \multirow{3}{*}{$\Biggl\{$} & EIL \cite{EIL}                     & 1.931  & 0.285  & 0.522 & 2.167   & 0.227   & 0.330  \\
                                                     &                                          & SPA \cite{SPA}                     & 5.528  & 0.221  & 0.357 & 7.425   & 0.169   & 0.262  \\
                                                     &                                          & TS-CAM \cite{TS-CAM}                  & 1.842  & 0.260  & 0.336 & 2.104   & 0.201   & 0.151  \\ \midrule
\multirow{5}{*}{\makecell{Weakly Supervised \\ Affordance Grounding}}  & \multirow{5}{*}{$\Vastt\{$} & Hotspots \cite{grounded}                & 1.773  & 0.278  & 0.615 & 1.994   & 0.237   & 0.577  \\
                                                     &                                          & Cross-view-AG \cite{ag_from_exocentric_imgs}           & 1.538  & 0.334  & 0.927 & 1.787   & 0.285   & 0.829  \\
                                                     &                                          & Cross-view-AG+ \cite{ag_from_exocentric_imgs+}          & 1.489  & 0.342  & 0.981 & 1.765  & 0.279   & 0.882  \\
                                                     &                                          & AffCorrs\dag \cite{affcorrs}                    & \underline{1.407} & \underline{0.359}  & \underline{1.026} & \underline{1.618}   & \underline{0.348}   & \underline{1.021} \\
                                                     &                                          & \method~ (Ours)                    & \textbf{1.226} & \textbf{0.401}  & \textbf{1.177} & \textbf{1.405}   & \textbf{0.372}   & \textbf{1.157} \\
\bottomrule
\end{tabular}
\caption{Comparison to state-of-the-arts from relevant tasks on AGD20K dataset. Results of * are taken from \cite{ag_from_exocentric_imgs}, and \dag~denotes the adapted AffCorrs. The \textbf{best} and \underline{second-best} results are highlighted in bold and underlined, respectively ($\uparrow$/$\downarrow$ means higher/lower is better).}
\label{tab:com_sota}
\end{table*}

\section{Experiments}

\subsection{Experimental Setting}
\noindent\textbf{Dataset and Metrics.} We evaluate our method in the Affordance Grounding Dataset (AGD20K)~ \cite{ag_from_exocentric_imgs}, which is the only large-scale image dataset with both exocentric and egocentric views.
AGD20K is comprised of 20,061 exocentric images and 3,755 egocentric images, and is annotated with 36 commonly used affordances.
Following prior affordance grounding work \cite{demo2vec, grounded}, the ground truth of this dataset initially consists of densely annotated points in corresponding affordance regions, and a Gaussian blur is then applied over each point to get final heatmaps.
Moreover, AGD20K can be evaluated in two different settings: 1) In the seen setting, object categories in training and test sets are identical.
2) In the unseen setting, there is no object category intersection between training and test sets, e.g., the model observes how humans hold a hammer and anticipates where to hold a knife.

As for the metrics, referring to previous affordance grounding work \cite{grounded, ag_from_exocentric_imgs, demo2vec, joint_hand_hotspot}, we adopt the commonly used Kullback-Leibler Divergence (KLD), Similarity (SIM), and Normalized Scanpath Saliency (NSS) to evaluate the similarity and correspondence of distributions between ground truth and prediction. Detailed calculation of each metric is shown in the supplementary material.

\vspace{1mm}
\noindent\textbf{Implementation Details.} 
We use the ImageNet \cite{imagenet} pretrained (without supervision) DINO-ViT-S \cite{dino-vit} with patch size 16 to generate deep features. 
In each iteration, N exocentric images along with one egocentric image are taken as input (N is set to 3). 
Images are first resized to $256\times256$ and then randomly cropped to $224\times224$ followed by random horizontal flipping.
SGD with learning rate 1e-3, weight decay 5e-4, and batch size 16 is used for parameter optimization.
Loss weight coefficients ($\lambda_{cos}, \lambda_{c}$) are set to (1, 0.07), and the margin $\alpha$ is set to 0.5.
For the first epoch, we warm up the network without $\mathcal{L}_{cos}$, as initial localization maps are not accurate for supervision.

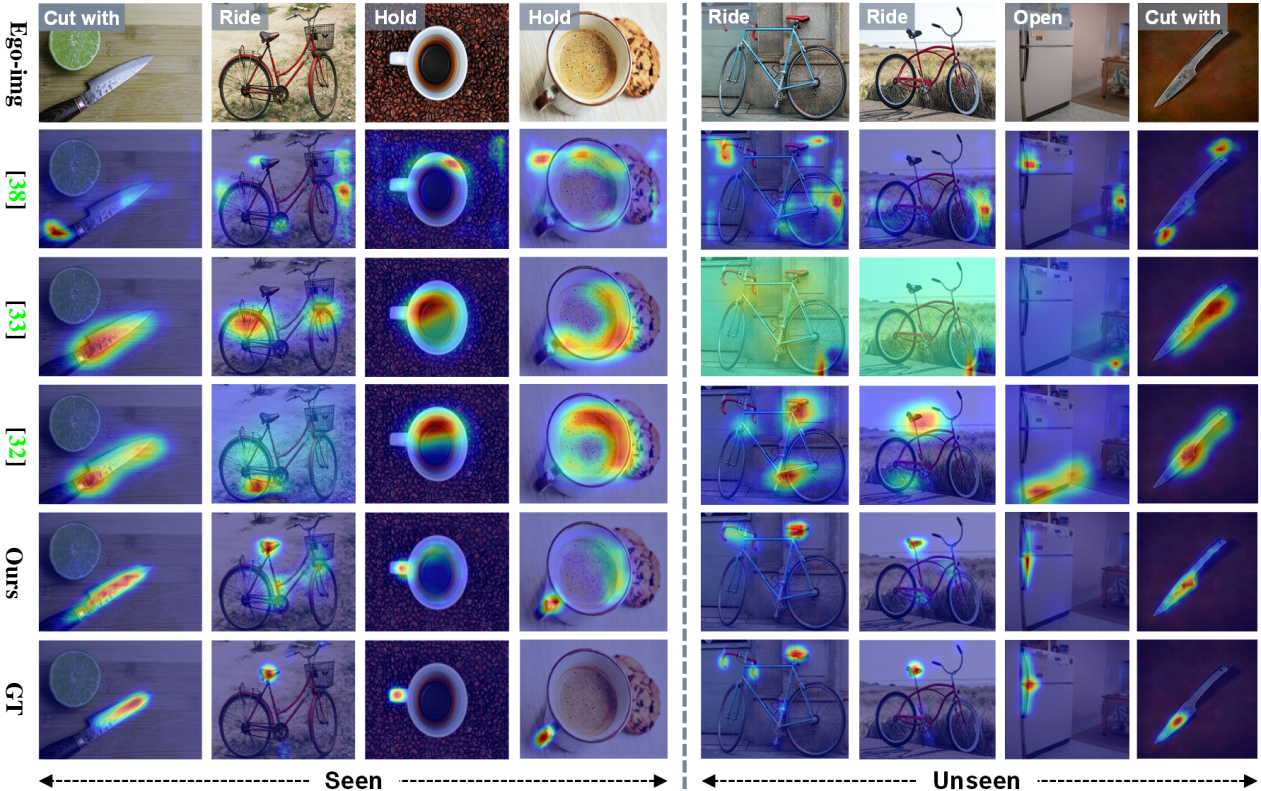
\begin{figure*}[!ht]
\centering
  \begin{overpic}[width=0.95\linewidth]{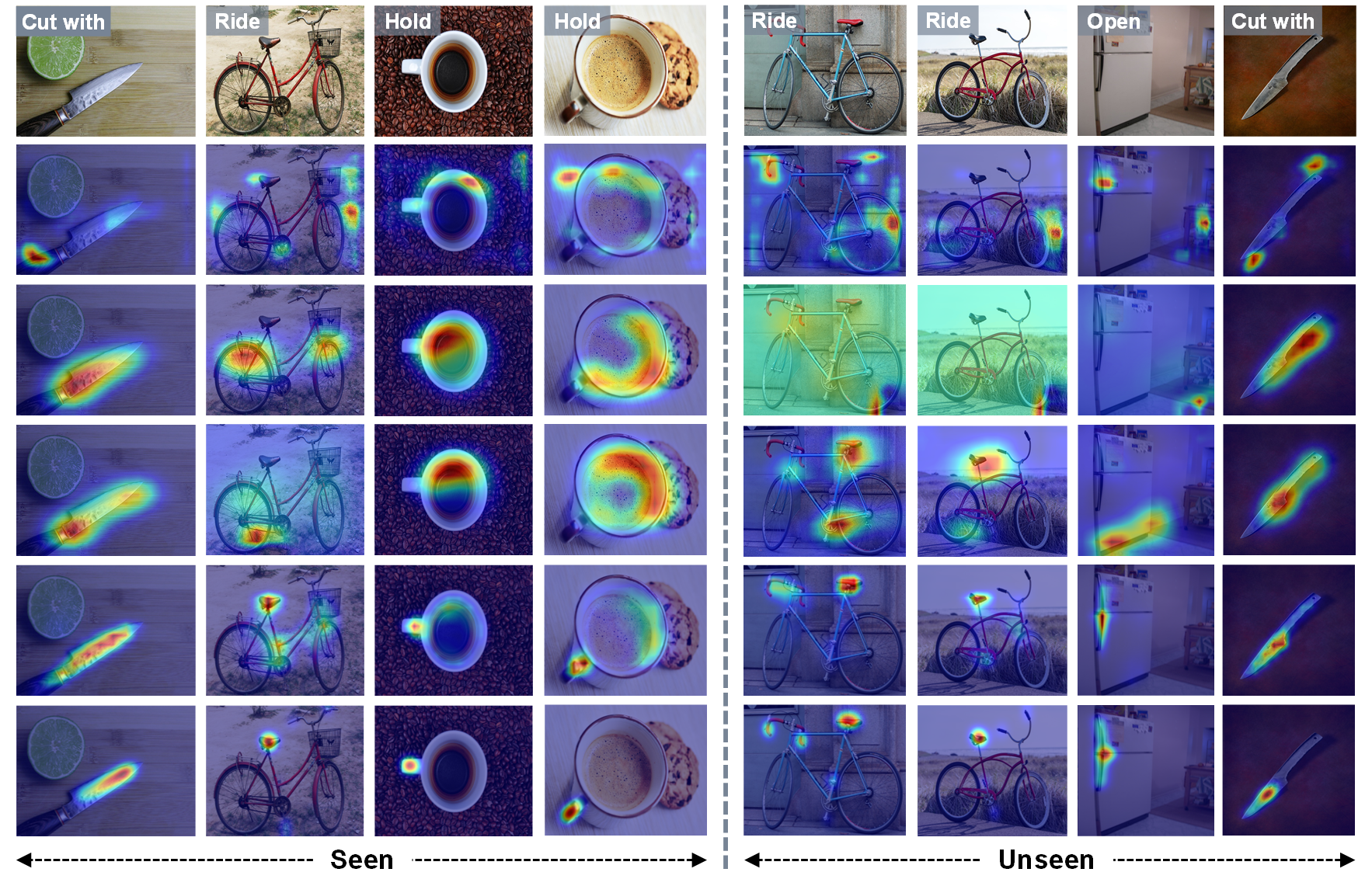}
        \put(-1.5,63.3){\rotatebox{-90}{\small\bfseries{Ego-img}}}
      \put(-1.5,51.8){\rotatebox{-90}{\small\bfseries{\cite{grounded}}}}
        \put(-1.5,41.4){\rotatebox{-90}{\small\bfseries{\cite{ag_from_exocentric_imgs}}}}
        \put(-1.5,31.2){\rotatebox{-90}{\small\bfseries{\cite{ag_from_exocentric_imgs+}}}}
        \put(-1.5,21.1){\rotatebox{-90}{\small\bfseries{Ours}}}
        \put(-1.5,10.3){\rotatebox{-90}{\small\bfseries{GT}}}

    \end{overpic}
  \caption{Qualitative comparison between our approach and state-of-the-art affordance grounding methods (Hotspots \cite{grounded}, Cross-view-AG \cite{ag_from_exocentric_imgs}, and Cross-view-AG+ \cite{ag_from_exocentric_imgs+}). For the unseen setting, the displayed objects are not in the training set. For example, the model learns where a \textit{motorcycle} can be ridden in training, and locates rideable area for the \textit{bicycle} at test time.
  }
  \label{fig:qua_com}
\end{figure*}

\subsection{Comparison to State-of-the-Art}
To conduct a comprehensive comparison, we also display the results of state-of-the-art methods from a relevant task, i.e., weakly supervised object localization.
As shown in Table~\ref{tab:com_sota}, in both seen and unseen settings, \method~ outperforms all other methods with a considerable margin on all metrics.
In particular, compared to the state-of-the-art affordance grounding method Cross-view-AG+ \cite{ag_from_exocentric_imgs+}, we improve the KLD by 20.4\%,
SIM by 33.3\%, and NSS by 31.2\% in the unseen setting.
Cross-view-AG+ is an extended version of Cross-view-AG, but still performs the knowledge transfer based on global pooled embeddings at the image level, thus bringing only minor improvement. 
AffCorrs \cite{affcorrs} is a method that focuses on one-shot part affordance grounding, and it also uses the pretrained DINO-ViT features to do part matching.
However, AffCorrs needs a pixel-level mask as a query, and there is no domain gap during the knowledge transfer.
To make AffCorrs comparable in our problem setting, we adapt its structure by replacing the query annotated mask with our CAM estimator.
The results verify that AffCorrs can also achieve good performance, but still considerably inferior to LOCATE.

\begin{table}[!t]
\centering
\begin{tabular}{lcc}
\toprule
Methods         & Params (M) & Time (s) \\ \midrule
EIL \cite{EIL}            & 42.41      & 0.019    \\
SPA \cite{SPA}            & 69.28      & 0.081    \\
TS-CAM \cite{TS-CAM}         & 85.86      & 0.023    \\ \midrule
Hotspots \cite{grounded}       & 132.64     & 0.087    \\
Cross-view-AG \cite{ag_from_exocentric_imgs}  & 120.03     & 0.023    \\
Cross-view-AG+ \cite{ag_from_exocentric_imgs+} & 82.27      & 0.022    \\
AffCorrs\dag \cite{affcorrs} & \textbf{6.50}      & 0.205    \\
\method~ (Ours) & \textbf{6.50} & \textbf{0.011}    \\\bottomrule
\end{tabular}
\caption{Comparison of learnable parameters and inference time. The inference time is evaluated on a 3090Ti GPU. \dag~denotes the adapted AffCorrs.}
\label{tab:param&time}
\end{table}

In Table~\ref{tab:param&time}, we make comparisons in terms of model parameters and inference time.
Since our framework is built on a frozen small-sized vision transformer (ViT-small), the training process is efficient with a small number of parameters.
For example, \method~ only has 5.4\% of learnable parameters in Cross-view-AG.
Additionally, we use a large patch size 16 for the vision transformer, which constrains the input sequence length and greatly reduces computation cost.
Therefore, the inference time of \method~ is also faster than most other methods.
By contrast, the adapted AffCorrs runs much slower than LOCATE, as it incorporates an additional CRF post-processing step.

We further visualize the qualitative comparisons with state-of-the-art affordance grounding methods.
As shown in Fig.~\ref{fig:qua_com}, we compare our results with Hotspots \cite{grounded}, Cross-view-AG \cite{ag_from_exocentric_imgs} and Cross-view-AG+ \cite{ag_from_exocentric_imgs+}.
We observe that the proposed \method~ can make more concentrated and accurate predictions.
Especially for complex objects like bicycles and refrigerators, even in the unseen setting, our method can still locate the saddle of bicycles for riding, and the handle of fridges for opening.
In comparison, the results of Cross-view-AG for bicycles are quite noisy.
More visualization results are in the supplementary material.

\subsection{Ablation Study}
\noindent \textbf{Knowledge Transfer Manner.}
We first investigate the impact of knowledge transfer manner. 
Previous affordance grounding methods \cite{grounded, ag_from_exocentric_imgs} simply pull close the global embeddings (produced by global average pooling) of two branches to perform global knowledge transfer (GKT).
In contrast, we set up an experiment to implement regional knowledge transfer (RKT), which generates the embeddings via masked average pooling between CAM-produced localization maps and feature maps.
The results are shown in Table~\ref{tab:ablation}, regional knowledge transfer (RKT) outperforms global knowledge transfer (GKT) on all metrics, demonstrating the effectiveness of filtering irrelevant information.

\begin{table}[!t]
\centering
\small
\begin{tabular}{@{}lcccccc@{}}
\toprule
\multirow{2.5}{*}{Method} & \multicolumn{3}{c}{Seen} & \multicolumn{3}{c}{Unseen} \\ \cmidrule(lr){2-4} \cmidrule(l){5-7} 
                        & KLD$\downarrow$    & SIM$\uparrow$    & NSS$\uparrow$    & KLD$\downarrow$     & SIM$\uparrow$     & NSS$\uparrow$    \\ \midrule
GKT                     &   1.732     & 0.267       &   0.810     &    1.971     &   0.221      &   0.626     \\ \midrule
RKT                     &   1.516     &     0.320   & 1.074       &     1.823    &    0.259     &  0.850      \\
+ $\mathcal{L}_{c}$                   &      1.491  &  0.326      & 1.091       &   1.750      &  0.274       & 0.948    \\
+ $\mathcal{S}$               &     1.236   & 0.397       &       1.178 &     1.439    &  0.358       & 1.130       \\
+ $\mathcal{S}$ + $\mathcal{L}_{c}$                &   1.226     & 0.401       &    1.177    &  1.405       &    0.372     & 1.157
\\ \bottomrule
\end{tabular}
\caption{Ablation results of the proposed \method~ framework. GKT/RKT means global/regional knowledge transfer. $\mathcal{S}$ denotes \s~ combined with $\mathcal{L}_{cos}$, and $\mathcal{L}_{c}$ is the concentration loss.}
\label{tab:ablation}
\end{table}

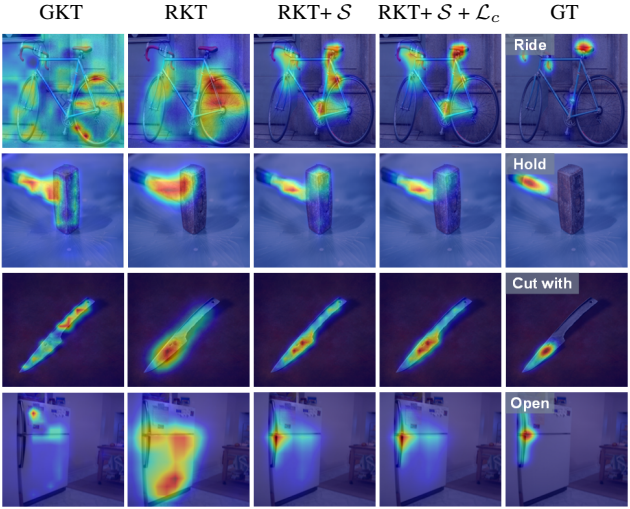
\begin{figure}[t]
\centering
  \begin{overpic}[width=1\linewidth]{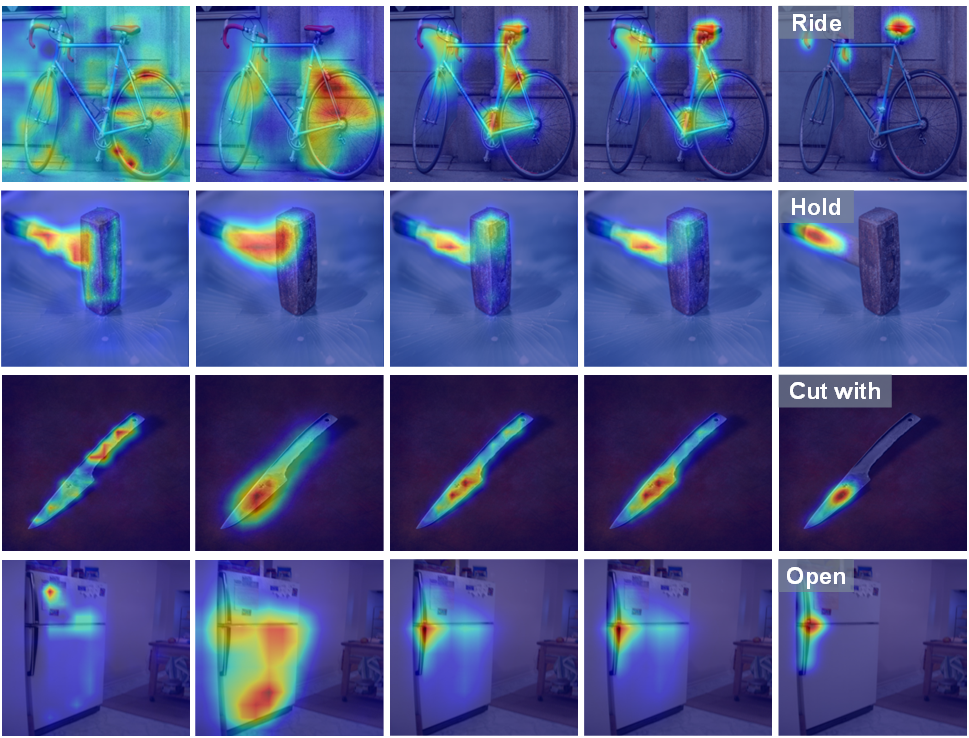}
        \put(6,78){{\footnotesize{GKT}}}
        \put(26,78){{\footnotesize{RKT}}}
        \put(44,78){{\footnotesize{RKT+ $\mathcal{S}$}}}
        \put(60,78){{\footnotesize{RKT}+ $\mathcal{S}$ + $\mathcal{L}_{c}$}}
        \put(87.5,78){{\footnotesize{GT}}}
    \end{overpic}
  \caption{Visualization of the qualitative improvements.}
  \label{fig:ablation_com}
\end{figure}

\vspace{1mm}
\noindent \textbf{\s~ and Concentration Loss.}
Based on the regional knowledge transfer, we analyze the effect of \s~ and concentration loss.
As shown in Table~\ref{tab:ablation}, directly applying the concentration loss $\mathcal{L}_{c}$ can only bring marginal improvement.
The reason is $\mathcal{L}_{c}$ can make egocentric predictions more concentrated, but fail to guide it to focus on the right affordance area.
Nonetheless, when adding \s~ and using cosine embedding loss $\mathcal{L}_{cos}$ as explicit supervision, the performance is greatly boosted, which proves the effectiveness of the part-level knowledge transfer scheme.
In addition, to check the qualitative improvement, we visualize the affordance grounding results in Fig.~\ref{fig:ablation_com}.
It is clear that GKT tends to locate the wrong affordance area, while RKT can sometimes find the right region, but only give coarse grounding results.
After adding \s~, the results become much more part-focused, and the concentration loss further makes the grounding maps more robust.

\begin{figure}[!ht]
\centering
    \begin{subfigure}[b]{\linewidth}
    \centering
        \includegraphics[width=1\linewidth]{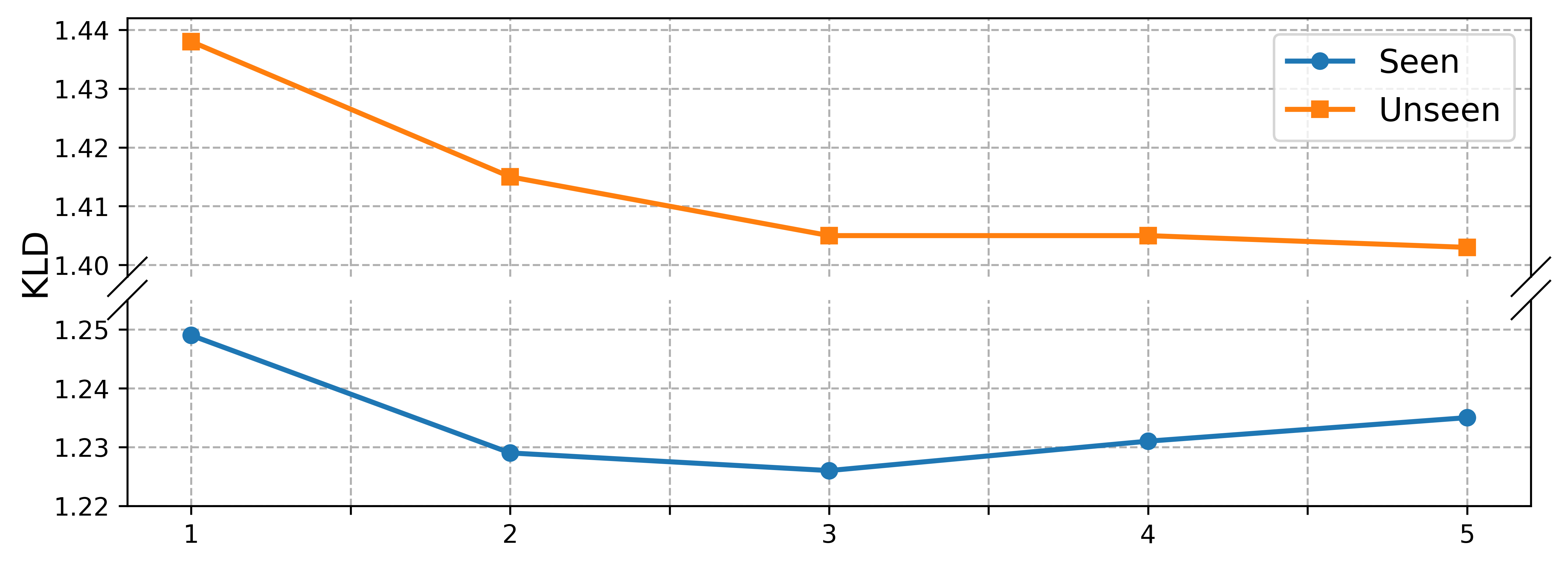}
        \caption{Number of prototypes $N$}
        \label{fig:fig_proto}
    \end{subfigure}
    
    \begin{subfigure}[b]{\linewidth}
    \centering
        \includegraphics[width=1\linewidth]{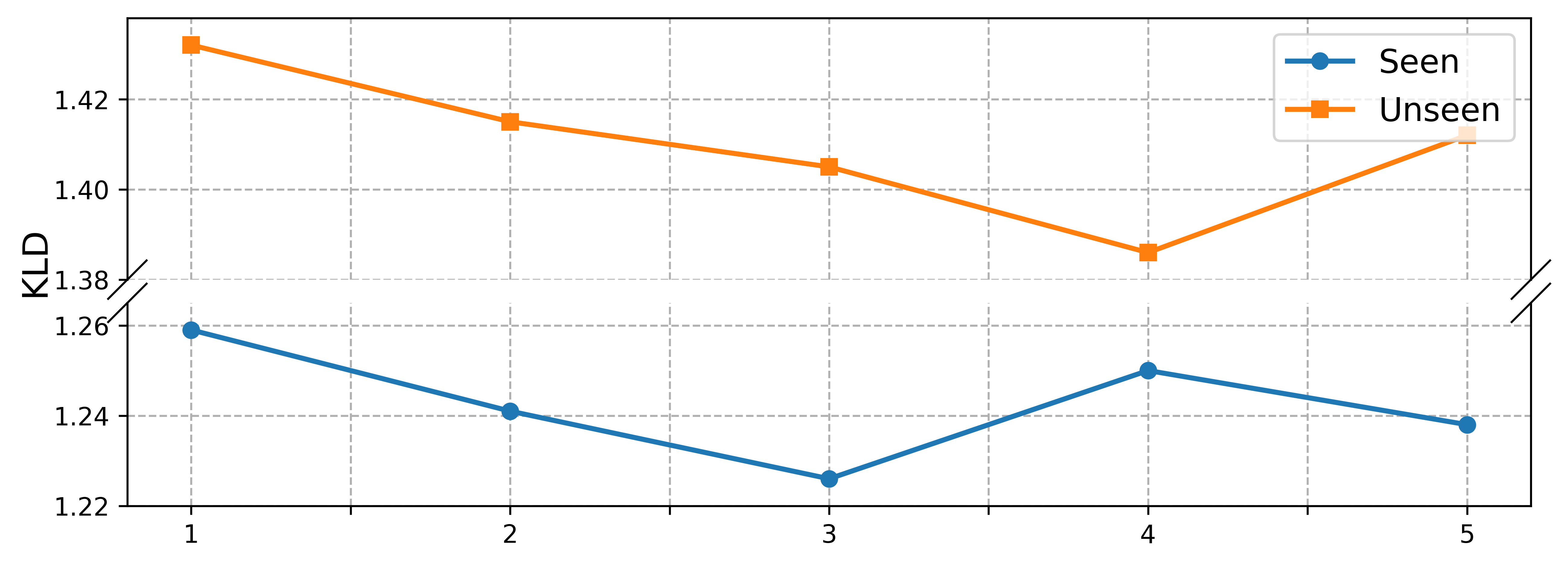}
        \caption{Number of exocentric images $K$}
        \label{fig:fig_exo}
    \end{subfigure}
\caption{Ablation study on the number of prototypes and exocentric images (lower is better).} 
\label{fig:num}
\end{figure}

\vspace{1mm}
\noindent \textbf{Number of Prototypes/Exocentric Images.}
We then explore the impact from the number of prototypes $K$ and exocentric images $N$.
From Fig.~\ref{fig:fig_proto}, we observe that the model yields the best performance with three prototypes in seen setting, which is consistent with our statement that interaction areas typically consist of human, object part, and background information.
While for the unseen setting, $N\!=\!5$ achieves the best results, but improvement is minor.
One reason lies in that more prototypes segment objects into more small parts, which boosts the generalization ability.
As for the number of exocentric images, we find that more exocentric images can alleviate the impact of interaction diversity and occlusion, thus providing more robust knowledge for the egocentric branch.
As shown in Fig.~\ref{fig:fig_exo}, the model gets largely improved when increasing the number of exocentric images from 1 to 3 in both seen and unseen settings.
Finally, we set both $N$ and $K$ to 3.

\begin{table}[!ht]
\centering
\small
\begin{tabular}{@{}rcccccc@{}} \toprule
                        & RN-50 & ViT-S/16 & \s~ & KLD$\downarrow$ & SIM$\uparrow$ & NSS$\uparrow$ \\
                        \cmidrule(r){1-4}
\cmidrule(l){5-7}
\multirow{4}{*}{\rotatebox[origin=c]{90}{Seen}}   & \checkmark     &     &          & 1.482    & 0.334    & 1.005    \\
                        &      & \checkmark    &          & 1.491    &  0.326   & 1.091    \\
                        &    \checkmark  &     &    \checkmark      &  1.449   & 0.340    &  1.021   \\
                        &      &  \checkmark   &   \checkmark       &  1.226   & 0.401    &  1.177   \\                         \cmidrule(r){1-4}
\cmidrule(l){5-7} 
\multirow{4}{*}{\rotatebox[origin=c]{90}{Unseen}} & \checkmark     &     &          & 1.701    & 0.287    & 0.962    \\
                        &      & \checkmark    &          &  1.750   &  0.274   & 0.948    \\
                        &  \checkmark    &     & \checkmark         & 1.707    & 0.287    & 0.949    \\
                        &      & \checkmark    & \checkmark         &  1.405   & 0.372    &  1.157  \\\bottomrule
\end{tabular}
\caption{Ablation study on different feature extractors.} 
\label{fig:feat}
\end{table}

\vspace{1mm}
\noindent \textbf{Different Feature Extractors.}
In LOCATE, we employ pretrained DINO features based on ViT, which have been proven to encode high-level semantic information \cite{deepvit}. 
To investigate the impact of different backbones, we conducted experiments with DINO features trained on the ResNet-50 \cite{resnet}.
From the results in Table \ref{fig:feat}, we observe that using DINO-ViT features directly (without the proposed \s~) can only obtain similar or even inferior results to their ResNet counterpart.
After incorporating \s~, the results of both backbone features can be improved under the seen setting, but ViT features show better potential in enhancing the performance due to their part-aware property.
In the unseen setting, \s~ does not yield improvement for ResNet-based features, while ViT features obtain consistent gains.

\section{Conclusion}
In this paper, we propose a framework named \method~ to address the weakly supervised affordance grounding task by observing human-object interaction images.
Specifically, we first localize where the interaction happens for the exocentric interactions, and then design a module called \s~ to pick the affordance-specific information from the interaction regions.
Finally, we transfer the learned knowledge to the egocentric view to perform affordance grounding with only image-level affordance labels.
The proposed \method~ achieves state-of-the-art results with far fewer parameters and faster inference speed.

\pagebreak  

{\small
\bibliographystyle{ieee_fullname}
\bibliography{egbib}
}

\end{document}